\documentclass[12pt,a4paper]{article}
\usepackage{url}
\usepackage{graphicx}
\usepackage{color}
\usepackage{alltt} % These for highlight

\begin{document}

\title{Still doing evolutionary algorithms with Perl}

\author{Juan Juli\'an Merelo Guerv\'os\\
Depto. Arquitectura y Tecnolog\'ia de Computadores, ETS Ingenier\'ia\\
  Inform\'atica y Telecomunicaciones, Universidad de Granada (Spain)\\
and\\
Granada Perl Monger\\
and\\
Free Software Office, UGR \url{http://osl.ugr.es}\\
Tel.: +34-958-243162,\\
Email: {\sf jmerelo@geneura.ugr.es}\\
\url{http://search.cpan.org/~jmerelo/}
}

\newcommand{\miae}{{\tt A::E}}

\maketitle

\begin{abstract}
{\sc Algorithm::Evolutionary} (\miae\ from now on) was introduced in 2002, after a talk in YAPC::EU in M\"unich. 7 years later,
\miae\ is in its 0.67 version (past its "number of
the beast" 0.666), and has been used extensively, to the point of
being the foundation of much of the (computer) science being done by
our research group (and, admittedly, not many others). 
All is not done, however; now \miae\ is being integrated with POE so
that evolutionary algorithms (EAs) can be combined with all kinds of servers
and used in client, servers, and anything in between. 
In this companion to the talk I will explain what evolutionary algorithms are,
what they are being used for, how to do them with Perl (using these or
other fine modules found in CPAN) and what evolutionary algorithms can
do for Perl at large.
\end{abstract}

\section{Don't talk no evolution, I believe in intelligent design!}
\label{sec:ae}

Right-on, buddy, but while you're at it think about this very simple
experiment. You've probably never done a {\em chili con carne}, right?
Well, if you {\em really} believe in {\em that} you might have.
OK, then go to your nearest hardware store and buy ten (yes, 10)
cast-iron pots, plus a notepad. Grab a ten-fire kitchen (that's a bit
more difficult, but, hey, this is a thought experiment, right?) and
put to cook ten {\em chili's con carne} with different ingredients. Here
you put a bit more {\em chili}, there a bit more {\em carne}, and up there a bit
more {\em con}. Note down carefully all you've done to all of them;
you'll end up with ten different recipes for chili con carne. Bring
down your (extended) family, and sit them down to eat, giving a score
to each one of your recipes. You'll end up with something like what we
show in table \ref{tab:score}.
\begin{table}
\caption{Scores given by Y. O. Ure family to the set of {\em chili con
  carne} recipes.\label{tab:score}}
\label{tab:ccc}       % Give a unique label
\begin{tabular}{l|c}
\hline\noalign{\smallskip}
Stew \# & Score  \\
\noalign{\smallskip}\hline\noalign{\smallskip}
0 & 43 \\
1 & 94 \\
2 & 3 \\
3 & 77 \\
4 & 82 \\
5 & 88 \\
6 & 21 \\
7 & 97 \\
8 & 41 \\
9 & 31 \\
\noalign{\smallskip}\hline
\end{tabular}
\end{table}
Well, right there you have pretty mean {\em chili's con carne}. And you
don't even need to do them all in a row, you can treat your (extended)
family 10 days in sequence until they stop remembering the good old
times when they ate something different to chili con carne. At any
rate, since you have the recipes for all of them, you decide to
convert \#2, \#6, and, for good measure, \#9 to biofuels, and try new
ones.

What can you do to make improvements in existing recipes for \#1 and \#7, which
were received with cheers and loud burps (actually, the score had to be
given in burps)? You pick a few quantities in the recipe for one of
them, and randomly mix it with the other. Let's say one of them was
a bit overcooked, and the other had a pinch more of cumin and secret
ingredient X\footnote{That would be cough syrup, but don't tell anyone},
you create recipe \#10 which is {\em both} overcooked, has
ingredient X and a cumin bonus track. It would be like breeding
boiling pots of chili, but without the pot and with extra
chili. That's one way of improving your already excellent recipe, but
you don't know in advance whether you're going to obtain a better result.

But there are more. While your aunt Fred (a particularly extravagant
member of your family nobody really wants to talk extensively about)
was tasting \#4, she said
\begin{verbatim}
This one is excellent, if only it had a bit more chorizo!
\end{verbatim}
She's partial to chorizo, this aunt Fred. Which is why you take the
recipe for number 4, and write down ``4 chorizo links'' instead of the
previous 2. That's a bit like when Peter Parker mutated into Spiderman
when bitten by radioactive spider, right? You insert a small
mutation in the recipe to create a new one.

And you do this over and over, you generate new recipes via small
changes and mixing what seems to be good (or simply feels right) of
old recipes, and, lo and behold, out comes intelligent life, sorry, a
whole family with a serious case of heartburn.

\section{Nature inspired optimization}

Natural evolution works a bit like that \cite{darwin2000osm}, with the
DNA acting as the chili recipes, and Nature itself working as your
family (yes, your aunt Fred too, never heard about platypus?). Only it
is a bit misleading to think that Nature actually optimizes species:
it adapts and builds on what already exists, but species are not in
the path to perfection same way as CPAN is not more perfect now than
it was ten years ago\footnote{Only it actually {\em is}}. 

However, searching the space of all possible chili recipes in this
fashion actually takes you to better and better recipes in time. Even
if not every time you make a change you obtain a better chili, the
whole set of stews improves each generation, by the simple procedure
of removing the worst and building on the best, the same way species
in Nature adapt and change, creating new ones. 

That is the basic idea behind evolutionary algorithms: incrementally
improving known solutions to problems by creating recipes for them (which might
be bit strings or more complicated data structures, like program trees), giving them an score (which
is called {\em fitness} in EA parlance), eliminate those with the
lowest fitness, and create new ones via changes ({\em mutation}) and
interchanges ({\em crossover}) of existing solutions. 

This kind of method was the one proposed by Holland \cite{Holland},
Hans-Paul Schwefel \cite{schwefel95} and eventually Goldberg
\cite{Goldberg}, who was one of the first that applied what were then
called {\em genetic algorithms} to engineering optimization
problems. Nowadays, evolutionary algorithms comprise a whole family of
algorithms including Genetic Programming  \cite{Koza90,Koza92} and Evolution
Strategies \cite{Hoffmeister90}, but their main differences are on the
specific representation they use for problem solutions (program trees,
floating point vectors, cellular automata) and the operators they
apply to them; all of them boil down (like chili) to the same pattern
of solutions improved (or not) via crossover and mutation in a
population.

\section{Evolving camels}
\label{sec:aeperl}

Despite this flexibility, there are not many libraries that implement evolutionary
algorithms in 
Perl (for a review on evolutionary algorithms in Perl, and a
tutorial of the first versions of {\tt A::E}, see
\cite{ecperl}); most published modules deal with genetic programming
(hereafter GP) in
Perl, due to the fact that it is an interpreted language, and it is very easy to
evaluate expressions and statements from a program (or script). The first
(as claimed by the author) paper published on the subject seems to be one by Baldi et
al. \cite{baldi00}, but the source code itself was not published, and
no hypothesis can be done on its features. From that moment on, there
are several papers that describe Perl implementation of evolutionary
algorithms: Kunken \cite{glotbot01} described an application
that evolves words that ``look as if they were English'', or {\em fake English} words, by trying to
evolve them using the same letter pattern that English uses. The same application is also mentioned by Zlatanov in
\cite{zlatanov01}, who implements a genetic programming system, with
source code available, to solve the same problem.

From then on, there are several papers about doing genetic programming
\cite{Koza92} in Perl: the 
first one was written by Murray and Williams \cite{murray99:ga}, which,
despite its title, actually describes a genetic programming system, similar to
another mentioned in the PerlMonks site \cite{gumpu01} (a meeting
place for practitioners). Several other introductions to genetic
algorithms with code have been published in the same place
\cite{evolve,improving:ea}, but
the first mention to a module that implements a canonical genetic
algorithm was done in \cite{ag01}. This module, called {\tt
  Algorithm::Genetic}, cannot be 
easily extended or adapted to new paradigms, since it is a single file
with all data structures and algorithms used already built-in into the
file. McCallum \cite{PerlGP} has also presented
a system called PerlGP, used specifically in the context of
bioinformatics, which has extensive facilities, including a database
back-end for serialization, and its main advantage is
that it uses as a programming language for doing GP in  Perl itself. Its main
drawback is its specificity: it is not intended for general evolutionary
computation, and most data structures and methods  are geared towards
GP. A later attempt is {\sf Algorithm::Evolve} \cite{algorithm:evolve}, which is well
designed, and quite easily extensible. Unfortunately, its development
stopped in 2003, and cannot be easily extended to include
representations different from the two default ones provided. {\sf
  Math::ES} \cite{math:es} comes approximately from the same date,
and is not intended as a general-purpose evolutionary computation
library, but rather designed for implementing the above mentioned
Evolution Strategies. This library if also frozen in the state it was
in 2003. 

The most complete (apart from the one presented in this paper) and
peculiar implementation of evolutionary 
algorithms in Perl was called {\sf myBeasties} \cite{hugman} and
eventually became a module called {\sf AI::GP}. This system
implements different kinds of objects, that can be evolved in many
possible ways; there's a language that describes these
transformations. It is an interesting system, but its extensibility is
not so strong, and the learning curve is also somewhat steep, since it
involves learning a new language apart from Perl itself. It is mainly
used for evolving Perl scripts, the same way that Genetic Programming
evolves Lisp functions, not intended for the implementation of a general evolutionary
computation program, which implies also learning structures unfamiliar
for the EA practitioner. 

On the other hand, one of the most recent is {\tt
AI::Genetic::Pro}, which has recently entered version 0.34. The main
objective of this module \cite{lukasz09:ai} is to optimize speed through
coding the most critical parts in C, through the Perl interface called
XS that allows this. In fact, initial tests\footnote{See equivalent programs at
our CVS server:
\url{http://opeal.cvs.sourceforge.net/viewvc/opeal/Algorithm-Evolutionary/benchmarks/},
{\tt ai-genetic-pro.pl} and {\tt bitflip.pl}} show that it is
several times slower than \miae, with extensibility being also
sacrificed through the use of this XS API. The other one is {\sf
  Math::Evol} \cite{math:evol}, which, as a differentiating trait, takes into account
constraints in search to guide evolution. It is mainly intended for straightforward
optimization problems, not as an extensible framework. The user has to
supply a set of problem-specific functions. However, this one is still
in development so who knows what its future will be. 

The majority of those systems do not make use Perl's capabilities to
implement an object-oriented library, easily adaptable and expandable,
which have been two of the objectives {\tt A::E}'s designers had in
mind. This, and the fact that the quality of the Perl programmer are
laziness, impatience and hubris, make up for the fact that I keep on
developing this library instead of paying some attention to the others
(which I should).

\section{Using \miae}
\label{sec:using}

Right behind making chili, one of the biggest problems in engineering
today is to place dots within sets of rectangles. It's probably not
very well known outside the rectangle engineering community, but
whenever you build a city in the clouds, optimally cover pole-dancing
poles within shoe boxes, or bake marshmallows, finding where the dot
has to {\em precisely} be is essential. But no worries, here's the
Perl program (split in Figures \ref{fig:program} and
\ref{fig:program:2}) that does exactly that by making use of the excellent
{\sf Algorithm::RectanglesContainingDot} module by Salvador Fandi\~no. 
\begin{figure}
\input {highlight.sty}
\noindent
\ttfamily
\small
\hlkwa{use\ }\hlstd{Algorithm}\hlsym{::}\hlstd{}\hlkwd{RectanglesContainingDot}\hlstd{}\hlsym{;}\hspace*{\fill}\\
\hlkwa{use\ }\hlstd{Time}\hlsym{::}\hlstd{HiRes\ }\hlkwd{qw}\hlstd{}\hlsym{(\ }\hlstd{gettimeofday\ tv\textunderscore interval}\hlsym{);}\hspace*{\fill}\\
\hlstd{}\hlkwa{use\ }\hlstd{Algorithm}\hlsym{::}\hlstd{Evolutionary\ }\hlkwd{qw}\hlstd{}\hlsym{(\ }\hlstd{Individual}\hlsym{::}\hlstd{BitString\ Op}\hlsym{::}\hlstd{Easy\hspace*{\fill}\\
}\hlstd{\ \ \ \ }\hlstd{Op}\hlsym{::}\hlstd{Bitflip\ Op}\hlsym{::}\hlstd{Crossover\ }\hlsym{);}\hspace*{\fill}\\
\hlkwc{my\ }\hlstd{}\hlkwb{\$alg\ }\hlstd{}\hlsym{=\ }\hlstd{Algorithm}\hlsym{::}\hlstd{RectanglesContainingDot}\hlsym{{-}$>$}\hlstd{}\hlkwd{new}\hlstd{}\hlsym{;}\hspace*{\fill}\\
\hlkwc{my\ }\hlstd{}\hlkwb{\$num\textunderscore rects\ }\hlstd{}\hlsym{=\ }\hlstd{shift\ }\hlsym{\textbar \textbar \ }\hlstd{}\hlnum{25}\hlstd{}\hlsym{;}\hspace*{\fill}\\
\hlstd{}\hlkwc{my\ }\hlstd{}\hlkwb{\$arena\textunderscore side\ }\hlstd{}\hlsym{=\ }\hlstd{shift\ }\hlsym{\textbar \textbar \ }\hlstd{}\hlnum{10}\hlstd{}\hlsym{;}\hspace*{\fill}\\
\hlstd{}\hlkwc{my\ }\hlstd{}\hlkwb{\$dot\textunderscore x\ }\hlstd{}\hlsym{=\ }\hlstd{shift\ }\hlsym{\textbar \textbar \ }\hlstd{}\hlnum{5}\hlstd{}\hlsym{;}\hspace*{\fill}\\
\hlstd{}\hlkwc{my\ }\hlstd{}\hlkwb{\$dot\textunderscore y\ }\hlstd{}\hlsym{=\ }\hlstd{shift\ }\hlsym{\textbar \textbar \ }\hlstd{}\hlnum{5}\hlstd{}\hlsym{;}\hspace*{\fill}\\
\hlkwc{my\ }\hlstd{}\hlkwb{\$bits\ }\hlstd{}\hlsym{=\ }\hlstd{shift\ }\hlsym{\textbar \textbar \ }\hlstd{}\hlnum{32}\hlstd{}\hlsym{;}\hspace*{\fill}\\
\hlstd{}\hlkwc{my\ }\hlstd{}\hlkwb{\$popSize\ }\hlstd{}\hlsym{=\ }\hlstd{shift\ }\hlsym{\textbar \textbar \ }\hlstd{}\hlnum{64}\hlstd{}\hlsym{;\ }\hlstd{}\hlslc{\#Population\ size}\hspace*{\fill}\\
\hlstd{}\hlkwc{my\ }\hlstd{}\hlkwb{\$numGens\ }\hlstd{}\hlsym{=\ }\hlstd{shift\ }\hlsym{\textbar \textbar \ }\hlstd{}\hlnum{50}\hlstd{}\hlsym{;\ }\hlstd{}\hlslc{\#Max\ number\ of\ generations}\hspace*{\fill}\\
\hlstd{}\hlkwc{my\ }\hlstd{}\hlkwb{\$selection\textunderscore rate\ }\hlstd{}\hlsym{=\ }\hlstd{shift\ }\hlsym{\textbar \textbar \ }\hlstd{}\hlnum{0.2}\hlstd{}\hlsym{;}\hspace*{\fill}\\
\hlslc{\#Generate\ random\ rectangles}\hspace*{\fill}\\
\hlstd{}\hlkwa{for\ }\hlstd{}\hlkwc{my\ }\hlstd{}\hlkwb{\$i\ }\hlstd{}\hlsym{(}\hlstd{}\hlnum{0\ }\hlstd{}\hlsym{..\ }\hlstd{}\hlkwb{\$num\textunderscore rects}\hlstd{}\hlsym{)\ \{}\hspace*{\fill}\\
\hlstd{\ \ }\hlstd{}\hlkwc{my\ }\hlstd{}\hlkwb{\$x\textunderscore 0\ }\hlstd{}\hlsym{=\ }\hlstd{}\hlkwd{rand}\hlstd{}\hlsym{(\ }\hlstd{}\hlkwb{\$arena\textunderscore side\ }\hlstd{}\hlsym{);}\hspace*{\fill}\\
\hlstd{}\hlstd{\ \ }\hlstd{}\hlkwc{my\ }\hlstd{}\hlkwb{\$y\textunderscore 0\ }\hlstd{}\hlsym{=\ }\hlstd{}\hlkwd{rand}\hlstd{}\hlsym{(\ }\hlstd{}\hlkwb{\$arena\textunderscore side}\hlstd{}\hlsym{);}\hspace*{\fill}\\
\hlstd{}\hlstd{\ \ }\hlstd{}\hlkwb{\$alg}\hlstd{}\hlsym{{-}$>$}\hlstd{}\hlkwd{add\textunderscore rectangle}\hlstd{}\hlsym{(}\hlstd{}\hlstr{"rectangle\textunderscore \$i"}\hlstd{}\hlsym{,\ }\hlstd{}\hlkwb{\$x\textunderscore 0}\hlstd{}\hlsym{,\ }\hlstd{}\hlkwb{\$y\textunderscore 0}\hlstd{}\hlsym{,\ }\hlstd{}\hlkwb{\$x\textunderscore 0}\hlstd{}\hlsym{+}\hlstd{}\hlkwb{\$side\textunderscore x}\hlstd{}\hlsym{,\ }\hlstd{}\hlkwb{\$y\textunderscore 0}\hlstd{}\hlsym{+}\hlstd{}\hlkwb{\$side\textunderscore y\ }\hlstd{}\hlsym{);}\hspace*{\fill}\\
\hlstd{}\hlsym{\}}\hspace*{\fill}\\
\hlslc{\#Declare\ fitness\ function}\hspace*{\fill}\\
\hlstd{}\hlkwc{my\ }\hlstd{}\hlkwb{\$fitness\ }\hlstd{}\hlsym{=\ }\hlstd{}\hlkwa{sub\ }\hlstd{}\hlsym{\{}\hspace*{\fill}\\
\hlstd{}\hlstd{\ \ }\hlstd{}\hlkwc{my\ }\hlstd{}\hlkwb{\$individual\ }\hlstd{}\hlsym{=\ }\hlstd{}\hlkwd{shift}\hlstd{}\hlsym{;}\hspace*{\fill}\\
\hlstd{}\hlstd{\ \ }\hlstd{}\hlkwc{my\ }\hlstd{}\hlsym{(\ }\hlstd{}\hlkwb{\$dot\textunderscore x}\hlstd{}\hlsym{,\ }\hlstd{}\hlkwb{\$dot\textunderscore y\ }\hlstd{}\hlsym{)\ =\ }\hlstd{}\hlkwb{\$individual}\hlstd{}\hlsym{{-}$>$}\hlstd{}\hlkwd{decode}\hlstd{}\hlsym{(}\hlstd{}\hlkwb{\$bits}\hlstd{}\hlsym{/}\hlstd{}\hlnum{2}\hlstd{}\hlsym{,}\hlstd{}\hlnum{0}\hlstd{}\hlsym{,\ }\hlstd{}\hlkwb{\$arena\textunderscore side}\hlstd{}\hlsym{);}\hspace*{\fill}\\
\hlstd{}\hlstd{\ \ }\hlstd{}\hlkwc{my\ }\hlstd{}\hlkwb{@contained\textunderscore in\ }\hlstd{}\hlsym{=\ }\hlstd{}\hlkwb{\$alg}\hlstd{}\hlsym{{-}$>$}\hlstd{}\hlkwd{rectangles\textunderscore containing\textunderscore dot}\hlstd{}\hlsym{(}\hlstd{}\hlkwb{\$dot\textunderscore x}\hlstd{}\hlsym{,\ }\hlstd{}\hlkwb{\$dot\textunderscore y}\hlstd{}\hlsym{);}\hspace*{\fill}\\
\hlstd{}\hlstd{\ \ }\hlstd{}\hlkwa{return\ }\hlstd{scalar\ }\hlkwb{@contained\textunderscore in}\hlstd{}\hlsym{;}\hspace*{\fill}\\
\hlstd{}\hlsym{\};}\hspace*{\fill}\\
\hlstd{}\hlslc{\#Initial\ population}\hspace*{\fill}\\
\hlstd{}\hlkwc{my\ }\hlstd{}\hlkwb{@pop}\hlstd{}\hlsym{;}\hspace*{\fill}\\
\hlstd{}\hlkwa{for\ }\hlstd{}\hlsym{(\ }\hlstd{}\hlnum{0}\hlstd{}\hlsym{..}\hlstd{}\hlkwb{\$popSize\ }\hlstd{}\hlsym{)\ \{}\hspace*{\fill}\\
\hlstd{}\hlstd{\ \ }\hlstd{}\hlkwc{my\ }\hlstd{}\hlkwb{\$indi\ }\hlstd{}\hlsym{=\ }\hlstd{Algorithm}\hlsym{::}\hlstd{Evolutionary}\hlsym{::}\hlstd{Individual}\hlsym{::}\hlstd{BitString}\hlsym{{-}$>$}\hlstd{}\hlkwd{new}\hlstd{}\hlsym{(\ }\hlstd{}\hlkwb{\$bits\ }\hlstd{}\hlsym{);}\hspace*{\fill}\\
\hlstd{}\hlstd{\ \ }\hlstd{}\hlkwd{push}\hlstd{}\hlsym{(\ }\hlstd{}\hlkwb{@pop}\hlstd{}\hlsym{,\ }\hlstd{}\hlkwb{\$indi\ }\hlstd{}\hlsym{);}\hspace*{\fill}\\
\hlstd{}\hlsym{\}}\hspace*{\fill}\\
\mbox{}
\normalfont
\caption{{\tt find\_dot\_in\_rectangles.pl} tries to find the position
  where the dot would be inside a maximal amount of
  rectangles.  This program is available from our
  CVS server:
http://opeal.cvs.sourceforge.net/viewvc/opeal/Algorithm-Evolutionary/examples/find\_dot\_in\_rectangles.pl. Continues
in Figure \ref{fig:program:2}
}
\label{fig:program}
\end{figure}

\begin{figure}
\input {highlight.sty}
\noindent
\ttfamily
\small
\hlstd{}\hlslc{\#\ Variation\ operators}\hspace*{\fill}\\
\hlstd{}\hlkwc{my\ }\hlstd{}\hlkwb{\$m\ }\hlstd{}\hlsym{=\ }\hlstd{Algorithm}\hlsym{::}\hlstd{Evolutionary}\hlsym{::}\hlstd{Op}\hlsym{::}\hlstd{Bitflip}\hlsym{{-}$>$}\hlstd{}\hlkwd{new}\hlstd{}\hlsym{;\ }\hlstd{}\hlslc{\#\ Rate\ =\ 1}\hspace*{\fill}\\
\hlstd{}\hlkwc{my\ }\hlstd{}\hlkwb{\$c\ }\hlstd{}\hlsym{=\ }\hlstd{Algorithm}\hlsym{::}\hlstd{Evolutionary}\hlsym{::}\hlstd{Op}\hlsym{::}\hlstd{Crossover}\hlsym{{-}$>$}\hlstd{}\hlkwd{new}\hlstd{}\hlsym{(}\hlstd{}\hlnum{2}\hlstd{}\hlsym{,\ }\hlstd{}\hlnum{9\ }\hlstd{}\hlsym{);\ }\hlstd{}\hlslc{\#\ Rate\ =\ 9}\hspace*{\fill}\\
\hlstd{}\hlkwc{my\ }\hlstd{}\hlkwb{\$generation\ }\hlstd{}\hlsym{=\ }\hlstd{Algorithm}\hlsym{::}\hlstd{Evolutionary}\hlsym{::}\hlstd{Op}\hlsym{::}\hlstd{Easy}\hlsym{{-}$>$}\hlstd{}\hlkwd{new}\hlstd{}\hlsym{(\ }\hlstd{}\hlkwb{\$fitness\ }\hlstd{}\hlsym{,\ }\hlstd{}\hlkwb{\$selection\textunderscore rate\ }\hlstd{}\hlsym{,\ {[}}\hlstd{}\hlkwb{\$m}\hlstd{}\hlsym{,\ }\hlstd{}\hlkwb{\$c}\hlstd{}\hlsym{{]}\ )\ ;}\hspace*{\fill}\\
\hlstd{}\hlkwc{my\ }\hlstd{}\hlkwb{\$inicioTiempo\ }\hlstd{}\hlsym{=\ {[}}\hlstd{}\hlkwd{gettimeofday}\hlstd{}\hlsym{(){]};}\hspace*{\fill}\\
\hlstd{}\hlkwa{for\ }\hlstd{}\hlsym{(\ }\hlstd{}\hlkwb{@pop\ }\hlstd{}\hlsym{)\ \{}\hspace*{\fill}\\
\hlstd{}\hlstd{\ \ }\hlstd{}\hlkwa{if\ }\hlstd{}\hlsym{(\ !}\hlstd{defined\ }\hlkwb{\$\textunderscore }\hlstd{}\hlsym{{-}$>$}\hlstd{}\hlkwd{Fitness}\hlstd{}\hlsym{()\ )\ \{}\hspace*{\fill}\\
\hlstd{}\hlstd{\ \ \ \ }\hlstd{}\hlkwc{my\ }\hlstd{}\hlkwb{\$this\textunderscore fitness\ }\hlstd{}\hlsym{=\ }\hlstd{}\hlkwb{\$fitness}\hlstd{}\hlsym{{-}$>$(}\hlstd{}\hlkwb{\$\textunderscore }\hlstd{}\hlsym{);}\hspace*{\fill}\\
\hlstd{}\hlstd{\ \ \ \ }\hlstd{}\hlkwb{\$\textunderscore }\hlstd{}\hlsym{{-}$>$}\hlstd{}\hlkwd{Fitness}\hlstd{}\hlsym{(\ }\hlstd{}\hlkwb{\$this\textunderscore fitness\ }\hlstd{}\hlsym{);}\hspace*{\fill}\\
\hlstd{}\hlstd{\ \ }\hlstd{}\hlsym{\}}\hspace*{\fill}\\
\hlstd{}\hlsym{\}}\hspace*{\fill}\\
\hlslc{\#\ Start\ Evolutionary\ Algorithm}\hspace*{\fill}\\
\hlstd{}\hlkwc{my\ }\hlstd{}\hlkwb{\$contador}\hlstd{}\hlsym{=}\hlstd{}\hlnum{0}\hlstd{}\hlsym{;}\hspace*{\fill}\\
\hlstd{}\hlkwa{do\ }\hlstd{}\hlsym{\{}\hspace*{\fill}\\
\hlstd{}\hlstd{\ \ }\hlstd{}\hlkwb{\$generation}\hlstd{}\hlsym{{-}$>$}\hlstd{}\hlkwd{apply}\hlstd{}\hlsym{(\ }\hlstd{$\backslash$}\hlkwb{@pop\ }\hlstd{}\hlsym{);}\hspace*{\fill}\\
\hlstd{\ \ }\hlstd{}\hlkwc{print\ }\hlstd{}\hlstr{"\$contador\ :\ "}\hlstd{}\hlsym{,\ }\hlstd{}\hlkwb{\$pop}\hlstd{}\hlsym{{[}}\hlstd{}\hlnum{0}\hlstd{}\hlsym{{]}{-}$>$}\hlstd{}\hlkwd{asString}\hlstd{}\hlsym{(),\ }\hlstd{}\hlstr{"}\hlesc{$\backslash$n}\hlstr{"}\hlstd{\ }\hlsym{;}\hspace*{\fill}\\
\hlstd{}\hlstd{\ \ }\hlstd{}\hlkwb{\$contador}\hlstd{}\hlsym{++;}\hspace*{\fill}\\
\hlstd{}\hlsym{\}\ }\hlstd{}\hlkwa{while}\hlstd{}\hlsym{(\ (}\hlstd{}\hlkwb{\$contador\ }\hlstd{}\hlsym{$<$\ }\hlstd{}\hlkwb{\$numGens}\hlstd{}\hlsym{)}\hspace*{\fill}\\
\hlstd{}\hlstd{\ \ }\hlstd{}\hlsym{\&\&\ (}\hlstd{}\hlkwb{\$pop}\hlstd{}\hlsym{{[}}\hlstd{}\hlnum{0}\hlstd{}\hlsym{{]}{-}$>$}\hlstd{}\hlkwd{Fitness}\hlstd{}\hlsym{()\ $<$\ }\hlstd{}\hlkwb{\$num\textunderscore rects}\hlstd{}\hlsym{));}\hspace*{\fill}\\
\hspace*{\fill}\\
\hlstd{}\hlkwc{print\ }\hlstd{}\hlstr{"Best\ is:}\hlesc{$\backslash$n$\backslash$t\ }\hlstr{"}\hlstd{}\hlsym{,}\hlstd{}\hlkwb{\$pop}\hlstd{}\hlsym{{[}}\hlstd{}\hlnum{0}\hlstd{}\hlsym{{]}{-}$>$}\hlstd{}\hlkwd{asString}\hlstd{}\hlsym{(),}\hlstd{}\hlstr{"\ Fitness:\ "}\hlstd{}\hlsym{,}\hlstd{}\hlkwb{\$pop}\hlstd{}\hlsym{{[}}\hlstd{}\hlnum{0}\hlstd{}\hlsym{{]}{-}$>$}\hlstd{}\hlkwd{Fitness}\hlstd{}\hlsym{(),}\hlstd{}\hlstr{"}\hlesc{$\backslash$n}\hlstr{"}\hlstd{}\hlsym{;}\hspace*{\fill}\\
\hlkwc{print\ }\hlstd{}\hlstr{"}\hlesc{$\backslash$n$\backslash$n$\backslash$t}\hlstr{Time:\ "}\hlstd{}\hlsym{,\ }\hlstd{}\hlkwd{tv\textunderscore interval}\hlstd{}\hlsym{(\ }\hlstd{}\hlkwb{\$inicioTiempo\ }\hlstd{}\hlsym{)\ ,\ }\hlstd{}\hlstr{"}\hlesc{$\backslash$n}\hlstr{"}\hlstd{}\hlsym{;}\hspace*{\fill}\\
\mbox{}
\normalfont
\caption{(Continues
in Figure \ref{fig:program}) {\tt find\_dot\_in\_rectangles.pl} tries to find the position
  where the dot would be inside a maximal amount of
  rectangles (2nd part).  This program is available from our
  CVS server:
http://opeal.cvs.sourceforge.net/viewvc/opeal/Algorithm-Evolutionary/examples/find\_dot\_in\_rectangles.pl
}
\label{fig:program:2}
\end{figure}

This is a kind of minimal program to use an evolutionary
algorithm. The first part is devoted to generate a set of random
rectangles, and then a closure is declared as a fitness function,
which counts within how many rectangles the dot is in; it uses the
{\tt decode} function, which converts the binary representation the individual
member of the population into an array of numbers, which are then used
to compute how many rectangles contain those dots. Besides closures,
classes can also be used to implement fitness functions; in that case
the object must respond to the {\tt apply} method, returning the
fitness. In this case (as in many others) this function is the one
that will be maximized: we will generate a population of dots, and
evolve one that is in as many rectangles as possible.

The population we start from is generated next: it is a purely random
population, composed of bit-strings with randomly generated bits. 

After that, a couple of operators are generated (already in Figure
\ref{fig:program:2}; as said in section 
\ref{sec:ae}, we need two kind of operations: mutation (here declared
as {\tt \$m}) and crossover ({\tt \$c} here). The declaration of these
operators includes the frequency with which they will be
used. Priority, or {\em rate}, is 1 for mutation and 9 for crossover:
that means that 90\% of the new individuals will be generated by
crossover (combining the bitstrings of two individuals), and the rest
by mutation. Priorities are transformed to probabilities in runtime
whenever operators are applied, that way, operator rates can be changed
in runtime and new operators can easily be added to them. 

These operators are used to define the {\tt \$generation} object, which
processes a single generation, destroying the 10\% worst (that's the
0.1 in {\tt \$selection\_rate}), and substituting it by the offspring
of the rest. After that, the population is evaluated; this could be
done faster by {\tt map( \$\_->evaluate( \$fitness ), @pop )}.

The algorithm itself is a simple business: apply the generation object
until a dot in all rectangles is found (not very likely) or until the
maximum number of generations has been reached. This usually happens
in as few as 25 generations, but it might take a few more than
that. The whole business takes less than one second in my computer. 

\section{Now with POE}
\label{sec:poe}

As soon as you want to integrate an evolutionary algorithm with
anything else, from a standalone daemon to a existing web server, or
simulate parallel systems (which you might be interested in if you
want to know what happens when you split the population in two
islands, but don't care much about what's the actual time
improvement) it will be necessary to include it in an event loop
system such as POE \cite{caputo2003ppo}. Initially, we did this a bit
by hand \cite{jj:2008:PPSN}, but then we decided to create a POE {\em
  component} that handled evolutionary algorithms: {\sf
  POE::Component::Algorithm::Evolutionary}, which besides being one of
the CPAN modules with the longest name, handles genetic populations as
POE {\em sessions}, allowing them to proceed evolutively alongside
each other, and communicate in several possible ways. For instance,
the program shown in Figure \ref{fig:poe} would be a fraction of a program\footnote{Which is actually in
  \url{http://search.cpan.org/~jmerelo/POE-Component-Algorithm-Evolutionary-0.2.1/lib/POE/Component/Algorithm/Evolutionary/Island/POEtic.pm}} 
just like the one shown in Figure \ref{fig:poe}.
\begin{figure}
\input {highlight.sty}
\noindent
\ttfamily
\hlkwa{use\ }\hlstd{POE\ }\hlkwd{qw}\hlstd{}\hlsym{(}\hlstd{Component}\hlsym{::}\hlstd{Algorithm}\hlsym{::}\hlstd{Evolutionary}\hlsym{::}\hlstd{Island}\hlsym{::}\hlstd{POEtic}\hlsym{);}\hspace*{\fill}\\
\hlstd{}\hspace*{\fill}\\
\hlslc{\#Stuff\ here}\hspace*{\fill}\\
\hlkwc{my\ }\hlstd{}\hlkwb{\$generation\ }\hlstd{}\hlsym{=\ }\hlstd{Algorithm}\hlsym{::}\hlstd{Evolutionary}\hlsym{::}\hlstd{Op}\hlsym{::}\hlstd{CanonicalGA}\hlsym{{-}$>$}\hlstd{}\hlkwd{new}\hlstd{}\hlsym{(\ }\hlstd{}\hlkwb{\$rr\ }\hlstd{}\hlsym{,\ }\hlstd{}\hlkwb{\$selection\textunderscore rate\ }\hlstd{}\hlsym{,\ {[}}\hlstd{}\hlkwb{\$m}\hlstd{}\hlsym{,\ }\hlstd{}\hlkwb{\$c}\hlstd{}\hlsym{{]}\ )\ ;}\hspace*{\fill}\\
\hlstd{}\hlkwc{my\ }\hlstd{}\hlkwb{\$gterm\ }\hlstd{}\hlsym{=\ }\hlstd{new\ Algorithm}\hlsym{::}\hlstd{Evolutionary}\hlsym{::}\hlstd{Op}\hlsym{::}\hlstd{GenerationalTerm\ }\hlnum{10}\hlstd{}\hlsym{;}\hspace*{\fill}\\
\hlstd{}\hspace*{\fill}\\
\hlkwc{my\ }\hlstd{}\hlkwb{@nodes\ }\hlstd{}\hlsym{=\ }\hlstd{}\hlkwd{qw}\hlstd{}\hlsym{(\ }\hlstd{node\textunderscore 1\ node\textunderscore 2\ }\hlsym{);}\hspace*{\fill}\\
\hlstd{}\hlkwc{my\ }\hlstd{}\hlkwb{\%sessions}\hlstd{}\hlsym{;}\hspace*{\fill}\\
\hlstd{}\hlkwa{for\ }\hlstd{}\hlkwc{my\ }\hlstd{}\hlkwb{\$n\ }\hlstd{}\hlsym{(\ }\hlstd{}\hlkwb{@nodes\ }\hlstd{}\hlsym{)\{}\hspace*{\fill}\\
\hlstd{}\hlstd{\ \ }\hlstd{}\hlkwc{my\ }\hlstd{}\hlkwb{@nodes\textunderscore here\ }\hlstd{}\hlsym{=\ }\hlstd{}\hlkwd{grep}\hlstd{}\hlsym{(\ }\hlstd{}\hlkwb{\$\textunderscore \ }\hlstd{}\hlkwa{ne\ }\hlstd{}\hlkwb{\$n}\hlstd{}\hlsym{,\ }\hlstd{}\hlkwb{@nodes\ }\hlstd{}\hlsym{);}\hspace*{\fill}\\
\hlstd{}\hlstd{\ \ }\hlstd{}\hlkwb{\$sessions}\hlstd{}\hlsym{\{}\hlstd{}\hlkwb{\$n}\hlstd{}\hlsym{\}\ =}\hspace*{\fill}\\
\hlstd{}\hlstd{\ \ \ \ }\hlstd{POE}\hlsym{::}\hlstd{Component}\hlsym{::}\hlstd{Algorithm}\hlsym{::}\hlstd{Evolutionary}\hlsym{::}\hlstd{Island}\hlsym{::}\hlstd{POEtic\hspace*{\fill}\\
\ }\hlsym{{-}$>$}\hlstd{}\hlkwd{new}\hlstd{}\hlsym{(\ }\hlstd{Fitness\ }\hlsym{=$>$\ }\hlstd{}\hlkwb{\$rr}\hlstd{}\hlsym{,}\hspace*{\fill}\\
\hlstd{}\hlstd{\ \ \ \ \ \ \ \ }\hlstd{Creator\ }\hlsym{=$>$\ }\hlstd{}\hlkwb{\$creator}\hlstd{}\hlsym{,}\hspace*{\fill}\\
\hlstd{}\hlstd{\ \ \ \ \ \ \ \ }\hlstd{Single\textunderscore Step\ }\hlsym{=$>$\ }\hlstd{}\hlkwb{\$generation}\hlstd{}\hlsym{,}\hspace*{\fill}\\
\hlstd{}\hlstd{\ \ \ \ \ \ \ \ }\hlstd{Terminator\ }\hlsym{=$>$\ }\hlstd{}\hlkwb{\$gterm}\hlstd{}\hlsym{,}\hspace*{\fill}\\
\hlstd{}\hlstd{\ \ \ \ \ \ \ \ }\hlstd{Alias\ }\hlsym{=$>$\ }\hlstd{}\hlkwb{\$n}\hlstd{}\hlsym{,}\hspace*{\fill}\\
\hlstd{}\hlstd{\ \ \ \ \ \ \ \ }\hlstd{Peers\ }\hlsym{=$>$\ }\hlstd{$\backslash$}\hlkwb{@nodes\textunderscore here\ }\hlstd{}\hlsym{);}\hspace*{\fill}\\
\hlstd{}\hlsym{\}}\hspace*{\fill}\\
\hlstd{}\hspace*{\fill}\\
\hlkwb{\$poe\textunderscore kernel}\hlstd{}\hlsym{{-}$>$}\hlstd{}\hlkwd{run}\hlstd{}\hlsym{();}\hlstd{}\hspace*{\fill}\\
\mbox{}
\normalfont
\caption{Fraction of an evolutionary algorithm as a POE component.\label{fig:poe}}
\end{figure}

In Figure \ref{fig:poe} two sessions or {\em nodes} are created, each one of
them running a separate evolutionary algorithm. Nothing much seems to
happen here, but after each step of the algorithm, a single individual
is sent from one of the nodes to the other, in a island-hopping
way. This is the default way of operation of the {\em island model}
evolutionary algorithm \cite{braun1990sts}: each node is running its
own population and, from time to time, they interchange some
individuals. It might get a bit more complicated, depending on who you
send, who you chose to receive, and what you do with
them\footnote{Check, for instance, our interesting multikulti
  algorithm \cite{multikulti:cec09}, which sends the most different
  instead of the best}, but it boils down to that: islands, and a boat
to send things between them. For the time being, it uses POE's own
{\tt post} mechanism for posting (that is why it's called POEtic), but
more mechanisms are intended in the future, starting with SOAP and
following with anything else that can be easily integrated with it
(XMPP, anyone?)

\section{Don't ask what evolutionary algorithms can do for you}

Well, actually, you can. EAs can be used for search and optimization,
so you can search and optimize whatever you want. For instance, you
can search for the solution to the Mastermind game \cite{mastermind05} with the help of
the {\sf Games::Mastermind}, search for the best parameters to train a
neural net using {\sf MachineLearning::NeuralNetwork} or evolve
English-sounding words after analyzing text corpora using {\sf
  Text::NGrams} or suchlike. Perl is fast enough, even more so if you
try to optimize the interpreter for speed\footnote{A forthcoming paper
  compares the performance of this and other Java library, finding out
  that, precisely in this area, they are comparable and, in some
  cases, Perl might even be the winner} and if you try to optimize as
much as possible the areas where the application spends the most time:
fitness evaluation and application of evolutionary operators
(mutation, crossover). 

Of course, you can use it not only for playing, but also for research:
evolving a new data structure needs only 3 new classes: the one for
representing the data structure, which might even come for free if
it is amenable to be used inside {\sf
  Algorithm::Evolutionary::Individual::Any}, and a mutation and
crossover. You can use hashes, vectors, B-Trees, even hairier data
structures, provided you know how to change them incrementally and
combine them. Remember that each change need not be for the better: it
is the population that improves on average, not each individual, X-Men
style.

With respect to \miae\ itself, I will continue to develop it for the
foreseeable future; of course, some help will always be
appreciated. Maybe a bit more of profiling is needed to identify
bottlenecks; a bit has been done, but not in all possible
situations. Bugs are mostly under control, but I haven't tested for
coverage, so maybe some will arise in the future. Best practices
\cite{conway2005pbp} are also generally followed, but I'm not keen on
renaming variables or modules pre-2005 to this convention. It will
have to be done, eventually, I guess. 

Finally, I know there's no decent library out there without a shining
and singing GUI. This will be done eventually. Until then, no version
1.0.

\section*{Acknowledgements}

This paper has been funded in part by the Spanish MICYT projects NoHNES
(Spanish Ministerio de Educaci\'on y Ciencia - TIN2007-68083) and
TIN2008-06491-C04-01 and the 
Junta de Andaluc\'ia P06-TIC-02025 and P07-TIC-03044\footnote{These
  are only those that are current now, but there have been many more
  of them since 2002}. We are also grateful to those who
have contributed patches and code to {\tt A::E}, too
numerous to mention, but fully acknowledged in the general page for
the library (found at \url{http://fon.gs/ae-acks/})

\bibliographystyle{unsrt}      % basic style, author-year citations
\bibliography{ea-with-perl,perl,perlga,GA-general,geneura,EC-ref,gann,ae}

\end{document}